# Elucidating Meta-Structures of Noisy Labels in Semantic Segmentation by Deep Neural Networks

Yaoru Luo[1], Guole Liu[2], Yuanhao Guo[1] and Ge Yang[1,2*]

[1]National Laboratory of Pattern Recognition, Institute of Automation, Chinese Academy of Sciences, Beijing, China

[2]School of Artificial Intelligence, University of Chinese Academy of Sciences, Beijing, China

*Corresponding author(s). E-mail(s): ge.yang@ucas.edu.cn;

luoyaoru2019@ia.ac.cn; liuguole@ucas.edu.cn; yuanhao.guo@ia.ac.cn;

**Abstract**

Supervised training of deep neural networks (DNNs) by noisy labels has been studied extensively in image classification but much less in image segmentation. Our understanding of the learning behavior of DNNs trained by noisy segmentation labels remains limited. We address this deficiency in both binary segmentation of biological microscopy images and multi-class segmentation of natural images. We classify segmentation labels according to their noise transition matrices (NTMs) and compare performance of DNNs trained by different types of labels. When we randomly sample a small fraction (e.g., 10%) or flip a large fraction (e.g., 90%) of the ground-truth labels to train DNNs, their segmentation performance remains largely unchanged. This indicates that *DNNs learn structures hidden in labels rather than pixel-level labels per se in their supervised training for semantic segmentation*. We call these hidden structures "meta-structures". When labels with different perturbations to the meta-structures are used to train DNNs, their performance in feature extraction and segmentation degrades consistently. In contrast, addition of meta-structure information substantially improves performance of an unsupervised model in binary semantic segmentation. We formulate meta-structures mathematically as spatial density distributions. We show theoretically and experimentally how this formulation explains key observed learning behavior of DNNs.

## 1 Introduction

Deep neural networks (DNNs) have achieved remarkable success in challenging image segmentation tasks (Hesamian et al. 2019, Minaee et al. 2021). However, it is generally thought that their supervised training requires accurate pixel-level labels. Manual annotation of pixels is laborious. More importantly, it introduces label noise, especially in border regions of image objects. Despite that label noise is often more common in image segmentation than image classification, studies on the training of DNNs by noisy labels so far have focused largely on classification. Is it really necessary to accurately label each pixel of training images? How will segmentation performance of DNNs be influenced by different types of label noise? Answering these questions is critical to understanding the role of labels in training of DNNs. It will also provide important insights into the learning behavior of DNNs.

Different types of noisy labels have been used to study the learning behavior of DNNs in image classification, especially partially corrupted or randomly shuffled labels (Arpit et al. 2017, Zhang et al. 2021). In image segmentation, there are image-level and pixel-level label noise. Image-level label noise refers to erroneous semantic annotation of image objects, whereas pixel-level label noise refers to erroneous semantic annotation of image pixels. In this study we focus on pixel-level label noise because of its importance in practice.

Specifically, we examine the performance of DNNs trained by four different types of labels, summarized in Table 1 and illustrated in Figure 1. To quantitatively characterize the segmentation performance of DNNs trained by these labels, we experiment on two representative models, U-Net (Ronneberger et al. 2015) and DeepLabv3+ (Chen et al. 2018), using the same loss function (binary cross-entropy) and optimizer (stochastic gradient descent, SGD). Performance of DNNs trained by the labels ranks from the best to the worst as follows:

$$CL \approx RCL > PCL > RL. \qquad (1)$$

**Table 1** Different Types of Segmentation Labels

| Labels | Meaning | Description | Accurate Boundary? | Randomized Pixels? |
|--------|---------|-------------|--------------------|--------------------|
| **CL** | Clean Label | Ground truth labels | Yes | No |
| **RCL** | Randomized Clean Label | Randomly sampled or flipped pixel labels from ground truth | Yes | Yes |
| **PCL** | Perturbed Clean Label | Dilation/erosion/skeleton of ground truth | No | No |
| **RL** | Random Label | Randomly generated pixel labels | No | Yes |

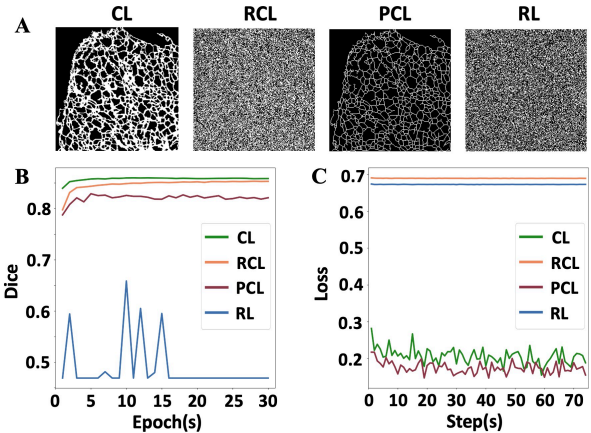

**Fig. 1** Segmentation performance of DNNs trained by different types of labels. (A) Four types of training labels for an image of the endoplasmic reticulum from the ER dataset (Luo et al. 2020). CL: ground truth from manual annotation. RCL: each pixel label in CL is randomly flipped with a probability of 0.45. PCL: Skeleton of CL. RL: each pixel is randomly labeled as 1, i.e., foreground, with a probability of 0.5. (B) Testing Dice scores during training. (C) Training loss of each optimization step.

As shown in Figure 1, when U-Net is trained with, for example, 45% of the labels randomly flipped (i.e., RCL) in binary segmentation, its performance remains largely the same as trained by the original ground truth (CL). Similar results are obtained on DeepLabv3+ (see Section 4). These results indicate that DNNs learn structures hidden in the noisy labels rather than the pixel labels *per se* in their training for segmentation. We call these hidden structures "meta-structures".

Similar as observed in image classification in (Zhang et al. 2021), we find that DNNs memorize random labels in segmentation because the training loss under RCL and RL quickly converges to a constant but not under CL and PCL (Figure 1C). Meanwhile, similar as observed in image classification in (Arpit et al. 2017), we find that before memorizing RL, DNNs prioritize learning real patterns in labels because the Dice score (Figure 1B, blue line) first fluctuates greatly then quickly drops to a low level. Motivated by the fact that RL requires no manual annotation, we develop a direct application of meta-structures in unsupervised learning for binary segmentation. Specifically, we develop a DNN model that sets RL as the initial training label and iteratively transits from RL to RCL by adding meta-structure information.

**Main Contributions**

A preliminary version (Luo et al. 2022) has been accepted by the 36th Association for the Advancement of Artificial Intelligence (AAAI-22) as an oral presentation. We extend it in five ways: (1) We characterize noisy labels formally using their noise transition matrices (NTMs) and show that they differ in the ranks of their NTMs. (2) To demonstrate that decrease in ranks of

NTMs leads to loss of semantic information, we quantify pixel-level randomization of noisy labels by defining a complete randomization distance. (3) To interpret how noisy labels degrades performance of DNNs, we visualize the feature patterns in hidden layers and the segmented errors. (4) We quantify the semantic information in different noisy labels and propose new theorems. (5) We conduct extensive experiments to validate the effectiveness of our unsupervised segmentation method.

Our contributions can be summarized as follows:

1. We provide direct experimental evidence that DNNs learn implicit structures hidden in noisy labels in semantic segmentation (Section 4). We refer to these implicit structures as meta-structures and model them mathematically as spatial density distributions of point patterns. We show theoretically and experimentally that this model explains key observed learning behavior of DNNs and that it quantifies semantic information in different segmentation labels (Section 7).

2. We have identified some fundamental properties of meta-structures and, therefore, noisy labels. We classify different labels based on their noise transition matrices (NTMs). We find that DNNs trained with labels under different perturbations to their meta-structures exhibit different yet consistently worse performance in feature extraction and segmentation (Section 5).

3. We demonstrate how to use meta-structure information in semantic segmentation. Specifically, by utilizing meta-structure information, we have developed an unsupervised model for binary segmentation that outperforms state-of-the-art unsupervised models and achieves remarkably competitive performance against supervised models (Section 6).

## 2 Related Work

***Performance of Models Trained by Noisy Labels in Image Classification.*** Previous studies have shown that DNNs can be trained to memorize large volumes of randomized labels but with poor generalization capability (Zhang et al. 2021) and that they prioritize learning real patterns of noisy labels before memorization (Arpit et al. 2017). These phenomena contradict traditional statistical learning theory (Vapnik 1992) and attract substantial research interest to characterize the generalization capability of DNNs trained by noisy labels. For example, it is shown empirically that generalization accuracy can be quantitatively characterized in terms of noise ratio in datasets (Chen et al. 2019). Upper bonds on generalization errors are also derived in terms of mutual information between input and output (Xu and Raginsky 2017). Furthermore, the information memorized by DNNs is quantified using Shannon mutual information between weights and labels (Harutyunyan et al. 2020). However, the focus of these studies is on image classification. In this study, we examine the performance of DNNs trained by different noisy labels in semantic segmentation.

***Methods to Handle Noisy Labels.*** In image classification, two types of methods are commonly used to deal with noisy labels: data cleansing and noise-tolerant enhancement. Data cleansing, which aims to filter out training labels that appear to be erroneous, reassigns labels by using alpha blending of given noisy labels (Jaehwan et al. 2019) or corrects labels using an ensemble of networks (Yuan et al. 2018). Suspicious labels can also be picked and sent to a human annotator for correction (Krause et al. 2016). Noise-tolerant enhancement, which aims to increase the robustness of learning systems under noisy training process, includes developing noise-robust loss functions (Masnadi-Shirazi and Vasconcelos 2008, Brooks 2011, Manwani and Sastry 2013, Van Rooyen et al. 2015, Ghosh et al. 2017, Zhang and Sabuncu 2018, Xu et al. 2019), designing custom architectures based on e.g., co-teaching (Han et al. 2018, Jiang et al. 2018), and utilizing multi-tasks frameworks to infer true labels (Veit et al. 2017, Tanaka et al. 2018, Li et al. 2020). Although most of these methods performed well in image classification tasks, their performance in image segmentation is unknown.

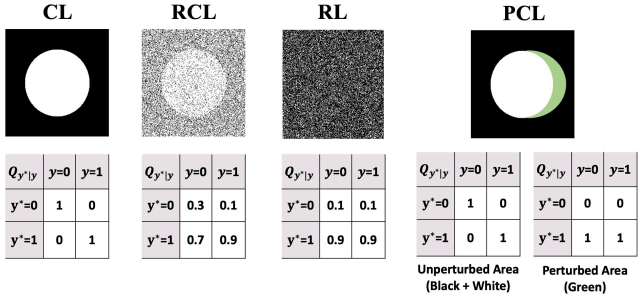

**Fig. 2** Examples of different types of binary semantic labels represented by their noise transition matrices (NTMs).

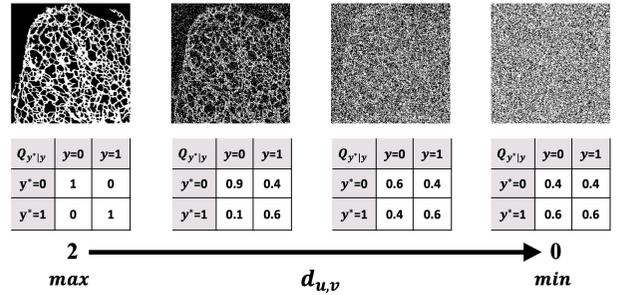

**Fig. 3** Examples of labels generated with different NTM. The randomization of pixel-level labels is gradually increasing from CL (left) to RL (right), as with the decrease of complete randomization distance $d_{u,v}$ from maximum (2) to minimum (0).

Unlike in image classification, much fewer studies on semantic segmentation focus on noisy labels. A mutual information-based approach is proposed to weaken the influence of back-propagated gradients caused by incorrect labels (Min et al. 2019). Local visual cues are leveraged to automatically correct label errors (Shu et al. 2019). Semi-supervised or unsupervised methods are also proposed (Navlakha et al. 2013, Lu et al. 2016, Li et al. 2019, Zheng and Yang 2021). Although these studies improved the generalization ability of DNNs trained with noisy labels, an in-depth understanding of the fundamental roles of segmentation labels in training remains lacking.

*Unsupervised Segmentation.* Some studies try to segment images by learning pixel representation in a self-supervised setting (Hwang et al. 2019, Zhang and Maire 2020). However, the proposed methods still rely on initialization from other annotated datasets. A small number of studies address image segmentation in a fully unsupervised way. For example, methods have been proposed to maximize the mutual information between augmented views (Ji et al. 2019, Ouali et al. 2020). A DNN architecture consisting of convolutional filters for feature extraction and differentiable processes for feature clustering has also been proposed (Kim et al. 2020). Overall, however, performance of unsupervised segmentation methods remains to be further improved.

## 3 Method

### 3.1 Definitions of different Labels

We define the four types of labels listed in Table 1 as follows:

*CL:* Clean labels (CLs) are ground-truth labels. They have accurate boundaries and non-randomized pixel labels.

*RCL:* Randomized clean labels (RCLs) are generated by randomly sampling with a probability $P_{sample}$ or flipping with a probability $P_{flip}$ pixel labels in CLs. For multi-class labels, a fraction of true pixel labels is randomly swapped with randomly selected labels from other classes. RCLs have accurate boundaries but randomized pixel labels. Examples of RCL are shown in Figure 4.

*PCL:* Perturbed clean labels (PCLs) are generated by dilation, erosion, or skeletonization of CLs. They have inaccurate object boundaries but non-randomized pixel labels. Examples of PCL are shown in Figure 7.

*RL:* Random labels (RLs) are generated by randomly assigning pixel labels with a probability $P_{generate}$. They can be considered as a strong perturbation of CLs because they contain no information from CLs. RLs have inaccurate boundaries and randomized pixel labels. Examples of RL are shown in Figure 9.

### 3.2 Representing Different Labels Using Noise Transition Matrices (NTMs)

Let $(\mathcal{X}, \mathcal{Y})$ denote an input image $\mathcal{X} \in R^{H \times W}$ with corresponding clean label image $\mathcal{Y} \in R^{H \times W}$, where $H$ and $W$ denote the height and width of the images, respectively. In $\mathcal{Y}$, each pixel label $y$ is correctly annotated by labels from a semantic class set $\{1, 2, \dots, M\}$. Based on the concept of class-conditional classification noise process (CNP) (Angluin and Laird 1988), which assumes correct labels are independently flipped to other classes, we use $p(y^* = j | y = i)$ to denote the probability of flipping pixel label $y$ in class $i$ to the noisy pixel label $y^*$ in class $j$. Using $p_{j|i}$ as a shorthand for flipping probability $p(y^* = j | y = i)$, the NTM $Q_{y^*|y}$ of label $\mathcal{Y}$ is constructed as a $M \times M$ matrix of $p_{j|i}$. That is, each column of $Q_{y^*|y}$ denotes the flipping probability distribution of $y$, and the element on the $j^{th}$ row $i^{th}$ column equals $p_{j|i}$.

Thus, CL can be synthesized by flipping $\mathcal{Y}$ with an identity matrix. RCL can be synthesized by flipping $\mathcal{Y}$ with a $Q_{y^*|y}$ whose rank ranges between 1 and $M$. PCL can be synthesized by flipping $\mathcal{Y}$ with a series of $Q_{y^*|y}$. Within unperturbed areas, $Q_{y^*|y}$ is an identity matrix. Within perturbed areas, the rank of $Q_{y^*|y}$ smaller than $M$. RL can be synthesized by flipping $\mathcal{Y}$ with arbitrary $Q_{y^*|y}$ whose rank equals 1. Examples of these four types of labels and their $Q_{y^*|y}$ in the case of binary label images are shown in Figure 2.

### 3.3 Complete Randomization Distance (CRD)

We define a complete randomization distance $d_{u,v}$ to measure the level of randomization between pixel labels of class $u$ and class $v$:

**Table 2** Performance of U-Net on CL and RCL of binary-class datasets

| Dataset | Training Labels | Noise Ratio | Dice (%) |
|---------|----------------|-------------|----------|
| ER | CL | 0 | **85.9** |
| | RCL | $P_{sample} = 0.1$ | 85.5 |
| | | $P_{sample} = 0.3$ | 85.6 |
| | | $P_{sample} = 0.5$ | 85.8 |
| | | $P_{flip} = 0.5$ | 85.9 |
| | | $P_{flip} = 0.7$ | 85.5 |
| | | $P_{flip} = 0.9$ | 85.2 |
| MITO | CL | 0 | **81.1** |
| | RCL | $P_{sample} = 0.2$ | 81.1 |
| | | $P_{sample} = 0.4$ | 80.9 |
| | | $P_{sample} = 0.8$ | 81.0 |
| | | $P_{flip} = 0.4$ | 80.7 |
| | | $P_{flip} = 0.6$ | 80.7 |
| | | $P_{flip} = 0.8$ | 80.5 |

**Table 3** Performance of DeepLabv3+ on CL and RCL of multi-class datasets

| Dataset | Training Labels | Noise Ratio | IoU (%) |
|---------|----------------|-------------|---------|
| Cityscapes | CL | 0 | **64.8** |
| | RCL | $P_{sample} = 0.1$ | 64.6 |
| | | $P_{sample} = 0.3$ | 64.5 |
| | | $P_{sample} = 0.5$ | 64.8 |
| | | $P_{flip} = 0.5$ | 64.7 |
| | | $P_{flip} = 0.7$ | 64.6 |
| | | $P_{flip} = 0.9$ | 64.7 |
| Cityscapes-3 | CL | 0 | **94.3** |
| | RCL | $P_{sample} = 0.2$ | 94.3 |
| | | $P_{sample} = 0.4$ | 94.1 |
| | | $P_{sample} = 0.8$ | 93.7 |
| | | $P_{flip} = 0.4$ | 93.8 |
| | | $P_{flip} = 0.6$ | 93.4 |
| | | $P_{flip} = 0.8$ | 93.1 |

$$d_{u,v} = \|p_{k|u}, p_{k|v}\|_1 = \sum_{k=1}^{M} |p_{k|u} - p_{k|v}|, \quad (2)$$

where $M$ denotes the number of classes. $p_{k|u}$ and $p_{k|v}$ are the probabilities in column $y = u$ and $y = v$ of $Q_{y^*|y}$, respectively. $\|\cdot\|_1$ is L1 norm. As shown in Figure 3, for binary label images, the maximum of $d_{u,v}$ equals 2, indicating a non-random mapping between two classes. With decreasing $d_{u,v}$, the level of randomization of pixel labels between two classes gradually increases until it reaches complete randomization when $d_{u,v} = 0$. That is, CRD characterizes how far away the mapping between two classes is from complete randomization. If the CRD is zero, the mapping between the two classes is entirely random.

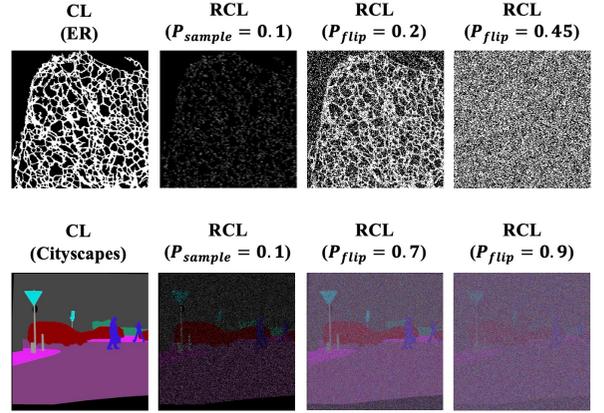

**Fig.4** Examples of RCL synthesized by random sampling/flipping with different probability ($P_{sample}/P_{flip}$).

### 3.4 Datasets and Experimental Configurations

We use two public fluorescence microscopy image datasets for binary-class semantic segmentation, including endoplasmic reticulum (ER) and mitochondria (MITO) (Luo et al. 2020). We use the Cityscapes (Cordts et al. 2016) dataset, which includes 19 urban scene classes, for multi-class semantic segmentation.

To simplify visualization and theoretical analysis of multiclass labels, we build another dataset by randomly choosing a subset of Cityscapes that contains only 3 of the 19 classes, for example the classes of road, sidewalk, and vegetation. We refer to this customized dataset as Cityscapes-3.

We choose U-Net (Ronneberger et al. 2015) for binary-class semantic segmentation because it is widely used in biomedical image segmentation. We choose DeepLabv3+ (Chen et al. 2018) for multi-class semantic segmentation because it is a benchmark model for natural image segmentation. Both models use SGD as the optimizer and cross entropy as the loss function. Detail hyperparameters for model training are shown in Supplementary Material A.2. The example images of each dataset are shown in Supplementary Material A.1. Code for reproducing results of this study is available at https://github.com/YaoruLuo/Meta-Structures-for-DNN.

## 4 Experimental Evidence for Existence of Meta-Structures

In this section, we examine the learning behavior of DNNs trained by CL and RCL in both binary and multi-class segmentation.

### 4.1 Existence of Meta-Structures

As shown in Figure 4, we synthesize a series of RCL with different flipping probabilities $P_{flip}$ on images from two semantic class dataset ER and multi semantic class dataset Cityscapes, respectively. Note that random sampling can be considered as a specific form of random flipping that swaps non-background pixel labels to background with probability $1 - P_{sample}$.

Table 2 and Table 3 show the segmentation results of U-Net and DeepLabv3+ on the binary-class and multi-class segmentation datasets, respectively. The maximum gap measured in absolute difference between percentage scores is 1.2. Taken together, these results show that DNNs learn structures hidden in labels rather than pixel-level labels per se in their supervised training. Furthermore, we reason that semantic information contained in CL is completely or largely preserved in RCL.

**Table 4** Segmentation performance in Dice score on binary-class datasets

| Dataset | | ER | | | | MITO | | | |
|---|---|---|---|---|---|---|---|---|---|
| $P_{0\|1}$ | | 0.2 | 0.4 | 0.6 | 0.8 | 0.2 | 0.4 | 0.6 | 0.8 |
| $1 - P_{0\|1} \pm \varepsilon$ | | | | | | | | | |
| - $\varepsilon$ | $\varepsilon \geq 0.1 (d \geq 0.2)$ | EQ | EQ | EQ | EQ | EQ | EQ | EQ | EQ |
| | $\varepsilon = 0.05 (d = 0.1)$ | -0.6% | -0.7% | -0.9% | -0.7% | -0.4% | -0.4% | -0.7% | -0.5% |
| | $\varepsilon = 0.01 (d = 0.02)$ | -3.7% | -2.8% | -3.1% | -2.4% | -4.5% | -4.4% | -3.9% | -1.8% |
| $\varepsilon = 0 (d = 0)$ | | OF | OF | OF | OF | OF | OF | OF | OF |
| + $\varepsilon$ | $\varepsilon = 0.01 (d = 0.02)$ | -3.3% | -3.9% | -3.0% | -2.9% | -3.8% | -6.3% | -4.6% | -2.4% |
| | $\varepsilon = 0.05 (d = 0.1)$ | -0.7% | -0.6% | -0.7% | -0.9% | -0.2% | -0.5% | -0.6% | -0.6% |
| | $\varepsilon \geq 0.1 (d \geq 0.2)$ | EQ | EQ | EQ | EQ | EQ | EQ | EQ | EQ |
| Baseline (Dice) | | 85.9% | | | | 81.1% | | | |

\* EQ: Equality in Performance. OF: Overfitting.

**Table 5** Segmentation performance in IoU on Cityscapes-3 dataset

| Labels | Rank | $Q_{y^*\|y}$ | Minimum $d_{u,v}$ | IoU (%) |
|---|---|---|---|---|
| CL | 3 | $\begin{matrix} 1 & 0 & 0 \\ 0 & 1 & 0 \\ 0 & 0 & 1 \end{matrix}$ | $d_{arb} = 2$ | **94.3** (Baseline) |
| RCL | 3 | $\begin{matrix} 0.5 & 0.3 & 0.2 \\ 0.2 & 0.6 & 0.2 \\ 0.3 & 0.1 & 0.6 \end{matrix}$ | $d_{1,3} = 0.6$ | 93.8 (-0.5) |
| | 3 | $\begin{matrix} 0.4 & 0.3 & 0.3 \\ 0.3 & 0.4 & 0.3 \\ 0.3 & 0.3 & 0.4 \end{matrix}$ | $d_{arb} = 0.2$ | 92.9 (-1.4) |
| | 3 | $\begin{matrix} 0.2 & 0.7 & 0.6 \\ 0.3 & 0.2 & 0.2 \\ 0.5 & 0.1 & 0.2 \end{matrix}$ | $d_{2,3} = 0.2$ | 92.7 (-1.6) |
| | 2 | $\begin{matrix} 0.5 & 0.5 & 0.2 \\ 0.2 & 0.2 & 0.2 \\ 0.3 & 0.3 & 0.6 \end{matrix}$ | $d_{1,2} = 0$ | 63.1 (-31.2) |
| | 2 | $\begin{matrix} 0.2 & 0.3 & 0.2 \\ 0.1 & 0.5 & 0.1 \\ 0.7 & 0.2 & 0.7 \end{matrix}$ | $d_{1,3} = 0$ | 36.9 (-57.4) |
| | 2 | $\begin{matrix} 0.6 & 0.3 & 0.3 \\ 0.1 & 0.6 & 0.6 \\ 0.3 & 0.1 & 0.1 \end{matrix}$ | $d_{2,3} = 0$ | 39.2 (-55.1) |
| | 1 | $\begin{matrix} 0.4 & 0.4 & 0.4 \\ 0.3 & 0.3 & 0.3 \\ 0.3 & 0.3 & 0.3 \end{matrix}$ | $d_{arb} = 0$ | OF |
| | 1 | $\begin{matrix} 0.1 & 0.1 & 0.1 \\ 0.2 & 0.2 & 0.2 \\ 0.7 & 0.7 & 0.7 \end{matrix}$ | $d_{arb} = 0$ | OF |
| | 1 | $\begin{matrix} 0.2 & 0.2 & 0.2 \\ 0.6 & 0.6 & 0.6 \\ 0.2 & 0.2 & 0.2 \end{matrix}$ | $d_{arb} = 0$ | OF |

\* $d_{arbitrary}$ Denotes complete randomization distance between arbitrary two classes. OF: Overfitting.

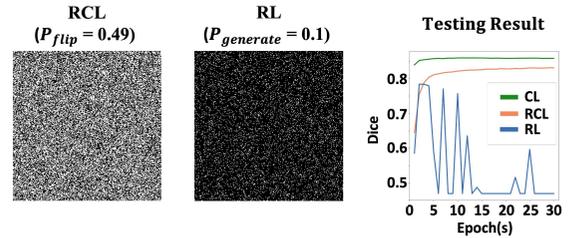

**Fig.5** Comparing training performance of RCL versus RL. Test dice scores of DNNs trained by CL, RCL ($P_{flip} = 0.49$) and RL ($P_{generate} = 0.1$) are compared in the rightmost panel.

## 4.2 Meta-Structures vs. Pixel-Level Labels

So far, we have shown that DNNs can learn from extremely noisy segmentation labels. However, it is unclear whether meta-structures or pixel-level labels contribute more to segmentation performance. For example, in binary-class segmentation when $P_{flip} \leq 0.49$, the fraction of correctly annotated pixels ($\geq 51\%$) still exceeds the fraction of incorrectly annotated pixels ($\leq 49\%$). This raises the possibility that DNNs learn from the majority of correct pixel-level labels rather than meta-structures.

To test this possibility, we generate entirely random labels, referred to as RL on the ER dataset. Each pixel is randomly assigned to foreground with a certain generation probability. A sample image is shown in Figure 5 (middle panel), where the generation probability is set as 0.1. In comparison, a RCL image is synthesized by random flipping with a probability $P_{flip} = 0.49$ (Figure 5, left panel). While the RCL contains the meta-structures, the RL does not. When we count the percentage of correctly annotated pixels using CL as the reference, we find that the pixel-level error rate of the RL is ~31%, which is much lower than the error rate of RCL (49%). If DNNs mainly learn from the pixel-level labels, the segmentation performance trained by RL would be better than by RCL.

However, segmentation performance of U-Net trained by RL is actually worse than by RCL (Figure. 5, right panel). This result further supports that DNNs learn from meta-structures in labels rather than pixel-level labels per se in their supervised training for semantic segmentation. Similar results of DeepLabv3+ are shown in Supplementary Material B.1.

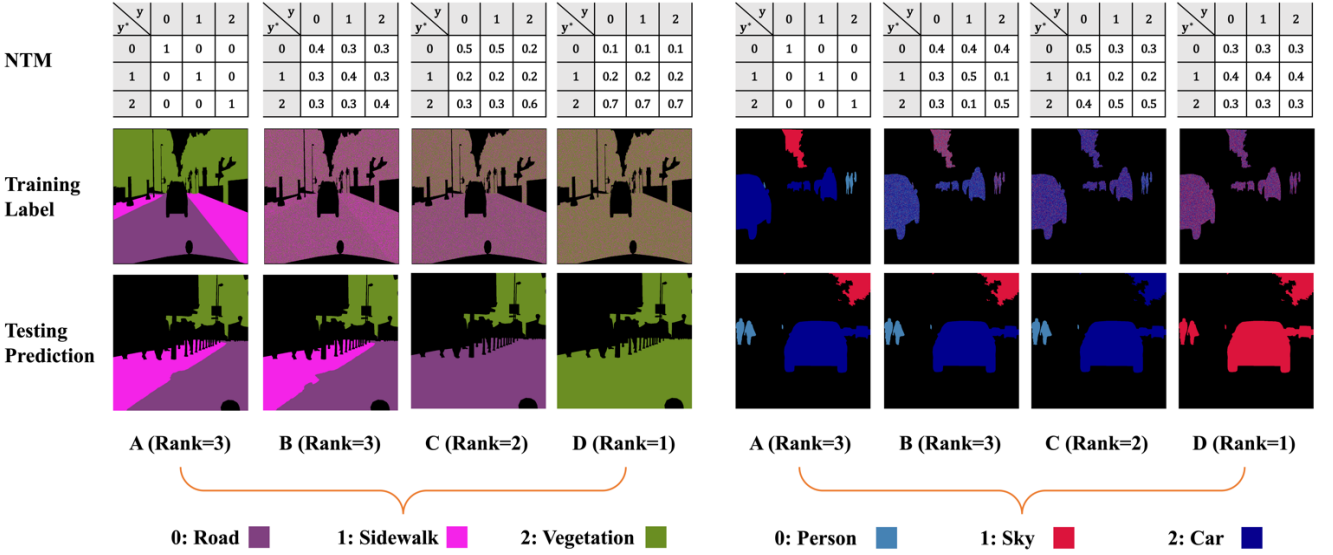

**Fig. 6** Testing results of DeepLabv3+ trained by CL (A) and RCL (B-D) on Cityscapes-3 dataset. Left: the select classes are Road, Sidewalk and Vegetation. Right: the selected classes are Person, Sky and Car. As the rank of NTM decreases, the number of correctly segmented classes decreases.

## 4.3 Negative Correlation between Segmentation Performance and CRD

Here we examine the relationship between the level of randomization of labels and the segmentation performance of DNNs trained by the labels. Specifically, we use CRD to quantify the level of randomization of labels.

*Binary Semantic Segmentation.* Based on $Q_{y^*|y} = \begin{bmatrix} 1 - P_{1|0} & P_{0|1} \\ P_{1|0} & 1 - P_{0|1} \end{bmatrix}$, we synthesized a series of RCL using different probability pairs $(P_{0|1}, P_{1|0} = 1 - P_{0|1} \pm \varepsilon)$, where $\varepsilon > 0$ and $P_{0|1} \in \{0.2, 0.4, 0.6, 0.8\}$. The CRD of RCL equals $2\varepsilon$. We use the result of CL as the baseline.

Table 4 shows the results of U-Net trained on the binary-class datasets. On ER dataset, we can clearly see a negative correlation between segmentation performance and CRD. Specifically, for each probability pair $(P_{0|1}, P_{1|0})$ with CRD $d = 0$, the model overfits. Surprisingly, however, even when $d$ is only slightly larger than 0, e.g., $d = 0.02$ when $Q_{y^*|y} = \begin{bmatrix} 0.79 & 0.8 \\ 0.21 & 0.2 \end{bmatrix}$, the model achieves nearly the same level of performance as the baseline, with the largest decrease of Dice scores measured in absolute percentage within 4%. Moreover, with slightly further increase CRD to $\varepsilon = 0.05$ ($d = 0.1$), the performance quickly recovers to nearly the same level as the baseline. When $\varepsilon \geq 0.1$ ($d \geq 0.2$), the performance trained by RCL completely matches the performance trained by CL.

Consistent results are obtained on the MITO dataset. In summary, DNNs exhibit strong robustness to pixel-level randomization noise in labels. However, their performance collapses when the labels are completely randomized.

*Multi-class Semantic Segmentation.* To simplify visualization of results on multi-class semantic segmentation, we experiment on the Cityscapes-3 dataset, which randomly select three semantic classes. We synthesize RCL with different $3 \times 3$ $Q_{y^*|y}$. DeepLabv3+ is used for training and testing. Table 5 shows the results. Figure 6 shows the visualization results trained with different noisy labels.

When RCL is generated under a NTM of rank 3, i.e., there are 3 classes with minimum CRD $d > 0$, the segmentation performance of DNNs is similar to DNNs trained by CL. All three semantic classes are correctly classified. Examples are shown in Figure 6 column B.

When RCL is generated under a NTM of rank 2, i.e. there exist two classes with CRD $d = 0$, the segmentation performance drops sharply and DNNs cannot distinguish between these two classes. Examples of segmented images are shown in Figure 6 column C.

When RCL is generated under a NTM of rank 1, i.e., $d = 0$ for any two classes, DNNs overfit and predict all semantic objects as one class. Examples of segmented images are shown in Figure 6 column D. Overall, DNNs exhibit similar behavior when trained with RL in multi-class semantic segmentation as in binary semantic segmentation.

## 4.4 Summary

Although RCL is uncommon in real-world applications, it reveals the counterintuitive learning behavior of DNNs that they learn meta-structures rather than pixel-level labels per se in segmentation. Specifically, results in Section 4.1 and Section 4.2 show that:

*DNNs trained by randomized labels that contain similar meta-structures as the ground truth labels provide similar performance in semantic segmentation.*

Furthermore, because CRD $d = 0$ will lead to degeneration of rank of $Q_{y^*|y}$, based on the experiments in section 4.3, we find that:

*The degeneration of rank of $Q_{y^*|y}$ will lead to loss of semantic information in labels.*

Mathematical formulation and proof of these two findings are presented later in Section 7.

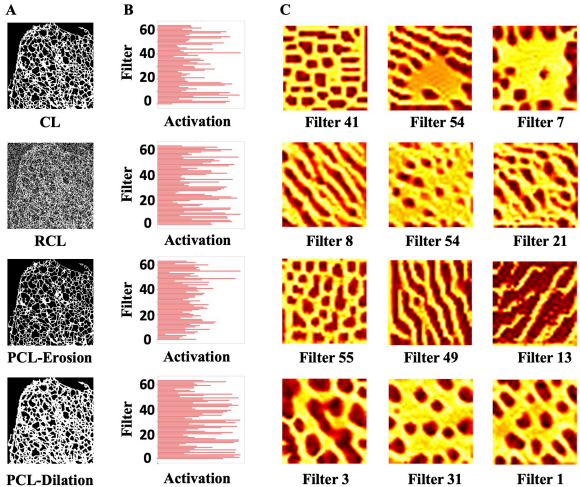

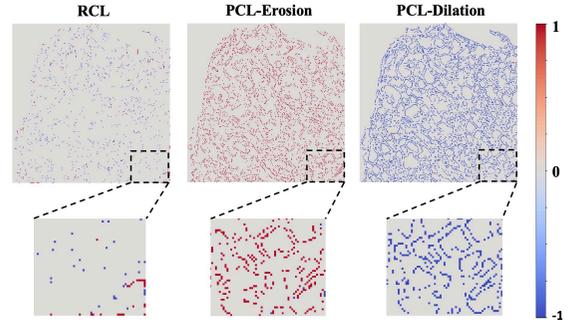

**Fig. 7** Feature patterns generated by different pre-trained DNNs. A: Different types of labels for training DNNs. B: Activation values of filters in the penultimate layer. C: Pattern images generated by top-3 filters according to Algorithm 1.

**Fig. 8** Difference images of segmented images between CL and noisy labels.

**Table 6** Segmentation performance of U-Net trained with PCL

| Dataset | Training Labels | Dice (%) |
|---------|-----------------|----------|
| ER | CL | **85.8** |
| | PCL-Erosion | 83.7 |
| | PCL-Dilation | 83.8 |
| MITO | CL | **80.9** |
| | PCL-Erosion | 75.4 |
| | PCL-Dilation | 78.7 |

## 5 Further Characterization of Meta-Structures

In Section 4 we have examined the learning behavior of DNNs trained by RCL. In this section we examine the learning behavior of DNNs trained by PCL and RL, respectively.

### 5.1 Learning Behavior of DNNs Trained by PCL

Inaccurate object boundaries are a common source of label noise in image segmentation. We simulate inaccurate boundaries using dilation and erosion of CL, which we refer to as PCL (Table 1). We examine the learning behavior of DNNs trained by PCL in the following two aspects: feature patterns and segmented images. Representative samples of PCL are shown in Figure 7A.

*Feature Patterns*. To elucidate how training with PCL affects the feature extraction of DNNs, we examine the feature patterns that DNNs prioritize recognizing.

The basic idea is that a pattern to which the convolutional filters of a DNN respond maximally could be a good first-order representation (Erhan et al. 2009). Based on this idea, we first feed the same image $I$ into different DNNs pre-trained by CL, RCL and PCL, respectively. Then we average the values of each feature map in the penultimate layer (i.e., before the softmax layer) to find the filters with the highest activation scores. This helps us find out the best filters for representing the input image $I$ of different pre-trained DNNs. Next, we use SGD to backwardly synthesize the pattern images that can maximally activate the selected filter by inputting a random image $X$. If noisy labels

---

**Algorithm 1 Pattern Synthesis Strategy**

**Require**: random image $X$, input image $I$, pre-trained model $\mathcal{F}$, hidden fitler $f$, number of layers $l$, interation steps $n$.

1: $f = argmax \ \mathcal{F}^{l-1}(I)$
2: **while** $t < n$ **do**
3:     min $\mathcal{L} = -f(X_t)$
3:     $X_{t+1} = X_t - \eta\nabla_X\mathcal{L}$
4:     $t = t + 1$
5: **end while**

---

(e.g., PCL) do not perturb feature extraction of DNNs, the generated pattern images should be consistent with the pattern images of DNNs trained by CL. The whole procedure is described in Algorithm 1.

We experiment on the ER dataset because we can easily identify structural features of the endoplasmic reticulum, specifically its thin tubules as well as wide sheets (magnified cartoon samples of tubules and sheets are shown in Figure 14). We pretrain U-Net by four types of labels as shown in Figure 7A, including CL, RCL, eroded PCL and dilated PCL. Figure 7B shows the activation of hidden filters in the penultimate layer (64 filters in total). Figure 7C shows the pattern images generated by the top three filters.

As shown in Figure 7, for CL and RCL, the pattern images include both tubule structures and sheet structures of endoplasmic reticulum. However, for PCL generated by erosion, the pattern structures tend to be thinner than those under CL and lack sheet objects. For PCL generated by dilation, the pattern structures tend to be thicker than those for CL. Thus, training DNNs with PCL will bias feature extraction.

*Segmentation Errors*. We use the difference image between segmentation results of DNNs trained by CL and corresponding segmentation results of DNNs trained by noisy labels to visualize segmentation errors under training by noisy labels. In binary segmentation result images, semantic objects and background are classified as 1 and 0, respectively. Thus, the difference image only contains three values, where 0 indicates matching results and $\pm 1$ indicates mismatching results. Representative difference images of RCL, erosion PCL and dilation PCL on the ER dataset are shown in Figure 8, respectively.

As shown in Figure 8, for the difference image of RCL, both $+1$ and $-1$ are sparsely distributed, indicating mostly random error but no systematic bias in the segmentation result under training by RCL. However, for erosion PCL and dilation PCL, the difference images show systematic bias near object boundaries, with mostly +1 for erosion PCL since it biases the segmentation result towards under-segmentation, and mostly −1 for dilation PCL since it biases the segmentation results towards over-segmentation. Table 6 summarizes results of U-Net trained by different PCLs on the binary-class datasets, which show consistent degradation in performance. Thus, training with PCL causes systematic bias in segmentation results.

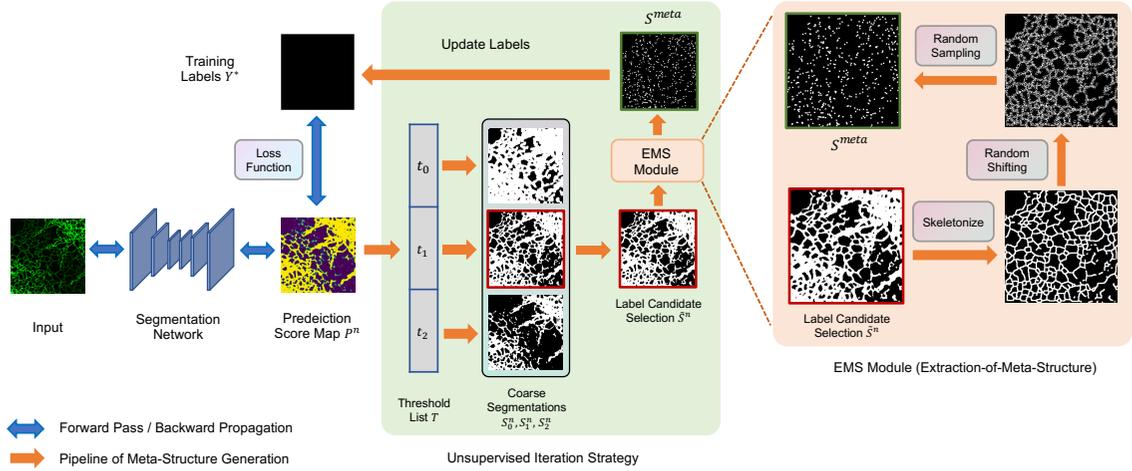

**Fig. 10** An overview of the unsupervised segmentation method iGTT. We first initialize the training labels with black images, then use $k$ thresholds to get $k$ separate coarse segmentation images and find the best candidate label $\tilde{S}^n$. Next, we use an EMS module to refine the meta structures of $\tilde{S}^n$ to $S^{meta}$. For EMS module(right), we first extract the skeleton of coarse label $\tilde{S}^n$, then randomly shift the pixels within a radius $r$, finally randomly sample the pixels by a probability $p_{sample}$. For visualization, the $K$, $r$, $p$ are set as 3, 2 and 0.1, respectively.

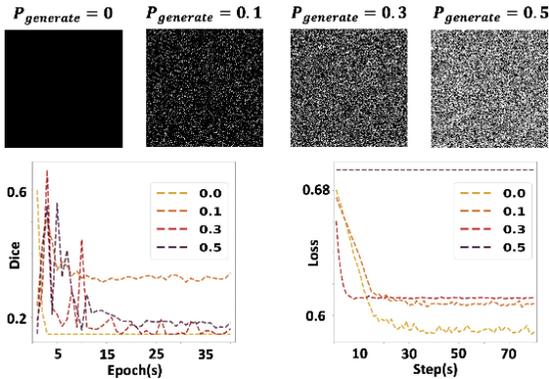

**Fig. 9** Testing dice score (left) and training loss (right) trained by RL.

### 5.2 Learning Behavior of DNNs Trained by RL

It has been shown that DNNs learn simple patterns before fitting RL by memorization (Arpit et al. 2017). This conclusion, however, is drawn in image classification. In our study, we investigate whether DNNs exhibit similar behavior in image segmentation. We generate RL under different generation probabilities (Figure 9) from 0 to 0.5. We experiment on the ER dataset and show the results of U-Net in Figure 9.

From the testing results on ER dataset (Figure 9, lower left panel), we find that the learning process consists of two stages. In the first stage, the Dice scores fluctuate substantially and reach several high values, indicating that the U-Net keeps learning and is not yet strongly influenced by RL. In the second stage, the Dice scores drop quickly then converge to a low value, indicating that the U-Net starts memorizing as its generalization ability becomes worse. Meanwhile, we find that under a higher generation probability, the training loss converges more quickly (Figure 9, lower right panel), indicating DNNs have a higher tendency to memorize random labels. Similar results are observed on DeepLabv3+. See Supplementary Material B.2 for further details. Overall, the learning behavior of DNNs trained by RL in segmentation is consistent with the observed learning behavior of DNNs trained by RL in classification.

### 5.3 Summary

Compared to CL and RCL, the PCL and RL contain different perturbations to meta-structures. We observe consistent degradation in segmentation performance and conclude that:

*Training of DNNs by labels with progressively stronger perturbation to the meta-structures exhibit progressively worse segmentation performance*

Mathematical formulation and proof of this argument is presented later in Section 7.

## 6 Unsupervised Binary Segmentation Using Meta-Structures

We propose an unsupervised model that requires no annotated labels for segmentation of images with two semantic classes. It is motivated by the observation that DNNs can still achieve good segmentation performance even when trained by RL in early phase of training, which provides a window of opportunity to transfer from RL to RCL by iteratively incorporating meta-structure information. We name our model iterative ground truth training (iGTT). Here, our goal is to show how to use meta-structures in practice rather than to develop a highly optimized binary segmentation model.

**Table 7** Segmentation performance of different models on ER dataset

| Training | Model | ER | | | | MITO | | | |
|---|---|---|---|---|---|---|---|---|---|
| | | DICE | AUC | ACC | IoU | DICE | AUC | ACC | IoU |
| Supervised | U-Net | 85.9 | 97.0 | 91.0 | 75.4 | 81.1 | 98.8 | 96.9 | 68.1 |
| | HRNet | 86.0 | 97.1 | 91.1 | 75.5 | 80.4 | 98.5 | 96.7 | 67.2 |
| | DeepLabv3+ | 81.6 | 94.8 | 87.6 | 69.1 | 78.1 | 98.1 | 96.3 | 64.1 |
| Unsupervised | AGT | 76.2 | 82.6 | 85.1 | 61.5 | 55.6 | 91.4 | 87.7 | 39.2 |
| | Otsu | 69.4 | 76.7 | 84.7 | 53.2 | 60.7 | 86.5 | 91.6 | 43.6 |
| | DFC | 78.1 | 85.2 | 84.4 | 63.8 | 54.1 | 84.6 | 80.7 | 38.4 |
| | AC | 73.1 | 87.8 | 81.4 | 57.6 | 66.2 | 94.8 | 93.8 | 49.5 |
| | iGTT (w EMS) | $78.8_{\pm 1.17}$ | $91.6_{\pm 1.04}$ | $85.4_{\pm 1.06}$ | $68.4_{\pm 0.96}$ | $70.9_{\pm 2.41}$ | $97.3_{\pm 0.67}$ | $94.7_{\pm 0.75}$ | $56.7_{\pm 2.15}$ |
| | iGTT (w/o EMS) | $73.9_{\pm 0.97}$ | $84.5_{\pm 2.52}$ | $81.1_{\pm 1.03}$ | $60.1_{\pm 1.26}$ | $68.1_{\pm 1.25}$ | $97.3_{\pm 0.19}$ | $94.2_{\pm 0.73}$ | $53.2_{\pm 1.83}$ |

* Numbers in subscripts represent standard deviation, calculated from 10 epochs.

---

**Algorithm 2 Unsupervised Iteration Strategy**

**Require**: input $X$, threshold $t$, model $\mathcal{F}$.

1: $n = 1$
2: $Y^* = 0$
3. **if** $n < Maxiters$ **then**
4.      $P^n = \mathcal{F}(X)$
5.      **UPDATE** $\mathcal{F}$ with $Y^*$
6.      **for** $k \in \{0, 1, ..., K-1\}$ **do**
7.          $S_k^n \leftarrow \begin{cases} p = 1 \ \textbf{if} \ p - t_k > 0, p \in P^n \\ p = 0 \ \textbf{if} \ p - t_k < 0, p \in P^n \end{cases}$
8.      **end for**
9.      $\tilde{S}^n = argmin_{k \in \{0,1,...,K-1\}} Cor(P^n, S_k^n)$
10.     $Y^* = S^{meta} = EMS(\tilde{S}^n)$
11.     $n = n + 1$
12: **end if**

## 6.1 Notation

Given image pair $(X, Y) \in \mathbb{R}^{H \times W}$, where $X$ denotes the input image, $Y$ denotes the binary label image with foreground and background labeled as 1 and 0, respectively. $H$ and $W$ denote image height and width, respectively. For iGTT, we use U-Net as the base model and only output the probability of each pixel belonging to the foreground. The output image is denoted as $P$.

## 6.2 Unsupervised Iteration Strategy

We use a fully black image to initialize the label image $Y^*$ then iteratively update it in following epochs. Specifically, for the $n_{th}$ epoch, we first update the model parameters $\theta$ based on current epoch label image $Y^*$. Then we directly threshold the current epoch predictions $P^n$ to generate $K$ coarsely segmented images $S_k^n$ ($k = 1, 2, ..., K$) from $K$ thresholds. The minimum threshold and maximum threshold are the minimum pixel value $P_{min}^n$ and maximum value $P_{max}^n$ of $P^n$, respectively. The interval of neighboring thresholds is $\frac{P_{max}^n - P_{min}^n}{K-1}$.

Next, based on an information-theoretic noise-robust loss $\mathcal{L}_{DMI}$ (Xu et al. 2019), we find the most correlated segmentation image $\tilde{S}^n$ that maximizes mutual information $Cor(P^n, S_k^n)$ between $P^n$ and $S_k^n$ ($k = 1, 2, ..., K$):

$$Cor(P^n, S_k^n) = \mathcal{L}_{DMI}(P^n, S_k^n) = -\log\left(\left|\det\left(Q_{(P^n \| S_k^n)}\right)\right|\right), \quad (3)$$

where $Q_{(P^n \| S_k^n)}$ is the matrix form of the joint distribution over $P^n$ and $S_k^n$. To calculate $Q_{(P^n \| S_k^n)}$, we resize $P^n \in \mathbb{R}^{H \times W}$ to $P_f \in \mathbb{R}^{1 \times HW}$, then concatenate $P_f$ and $1 - P_f$ to form $\mathcal{P} \in \mathbb{R}^{2 \times HW}$. Meanwhile, we resize $S_k^n \in \mathbb{R}^{H \times W}$ to $S_f \in \mathbb{R}^{1 \times HW}$, then concatenate $S_f$ and $1 - S_f$ to form $\mathcal{S} \in \mathbb{R}^{2 \times HW}$. Then $Q_{(P^n \| S_k^n)}$ is calculated by matrix multiplication $Q_{(P^n \| S_k^n)} = \mathcal{P}\mathcal{S}^T$.

The optimal $\tilde{S}^n$ that maximizes $Cor(P^n, S_k^n)$ is then sent into an EMS module (i.e., extraction-of-meta-structure module, see Section 6.3) to extract meta-structures $S^{meta}$. Finally, we update the label $Y^*$ by $S^{meta}$. The pseudo-code and the architecture of iGTT are shown in Algorithm 2 and Figure 10, respectively.

## 6.3 Extraction-of-Meta-Structure Module

Because of insufficient training in early training steps, the segmented images $\tilde{S}^n$ by thresholding are still rather coarse. However, the basic topology of objects is largely retained in $\tilde{S}^n$. Based on this, we design an extraction-of-meta-structure (EMS) module to further improve the quality of pseudo labels.

We first extract the skeleton of $\tilde{S}^n$, then we randomly shift every pixel in skeleton within a radius $r$ to approximate the width of meta-structures. Since the random shift may move some pixel labels outside the target meta-structures, we follow with a random sampling operation with probability $p_{sample}$ to filter out these pixel labels and generate the final pseudo label $S^{meta}$ for the next epoch of training.

## 6.4 Experiment Results

As iGTT is customized for binary-class segmentation, we evaluate its performance on the ER, MITO (Luo et al. 2020) datasets. The $K$, $r$ and $p_{sample}$ are set as 30, 1 and 0.1, respectively. We combine $\mathcal{L}_{DMI}$ and IoU as the loss function (Huang et al. 2019) to optimize our models.

To compare with supervised methods, we select U-Net, DeepLabv3+ and a state-of-the-art model HRNet (Wang et al. 2020). To compare with unsupervised methods, we select adaptive Gaussian Thresholding (AGT), Otsu, and two state-of-the-arts methods: Autoregressive Clustering (AC) (Ouali et al. 2020) and Differentiable Feature Clustering (DFC) (Kim et al. 2020). We use DICE (dice score), AUC (area under curve) and ACC (accuracy) as the performance metrics. To reduce the effects of randomization, we train our model 10 times and calculate the mean and standard deviation.

Segmentation results on binary-class datasets are summarized in Table 7. Among unsupervised methods, iGTT achieves the best performance. Meanwhile, we find that iGTT achieves competitive performance when compared with the other three supervise models. Moreover, we find that using EMS module improves the final segmentation performance, indicating that EMS indeed refines the candidate labels. The visualization of segmentation results between different models are shown in Figure 15.

Overall, by utilizing meta-structures in noisy labels, our model effectively narrows the gap between supervised learning and unsupervised learning.

# 7 Theoretical Modeling and Analysis of Meta-Structures

In this section, we model meta-structures of labels as spatial density distributions (SDD) of point patterns (Baddeley et al. 2015). Then we quantify the semantic information of different labels according to their meta-structures and use three theorems to further elucidate roles of training labels in segmentation.

## 7.1 Preliminaries

***Background on Spatial Density Distributions.*** Spatial density distributions are used to analyze implicit patterns of spatial point data. Kernel density estimation, a non-parametric method, estimates the spatial density of events within a specified region by using a kernel function to weight the area surrounding each point according to its distance to the event.

Specifically, for the spatial point $x$, its spatial density distributions $f(x)$ can be estimated as follows:

$$f(x) = \frac{1}{2h} \frac{\sum x_i \in [x-h, x+h]}{N} = \frac{1}{2Nh} \sum_{i=1}^{N} K\left(\frac{x-x_i}{h}\right), \quad (4)$$

where $N$ is the number of data points, $h$ is the bandwidth (radius) of a rectangle search window $W$, centered at $x$. $K$ is the kernel function.

***Notation.*** We use $M$ and $D$ to denote the number of semantic classes of uncorrupted image label $Y$ and noisy label $Y^*$, respectively. The noisy image labels $Y^*$ is synthesized by a $M \times M$ NTM $Q_{y^*|y}$. We treat the set of pixels $x^m$ whose label $y^* = m$ ($y^* \in Y^*$ and $m \in \{1, 2, \ldots, M\}$) as the spatial data point set by annotating them as 1 and treat other pixels $x^m$ ($y^* \neq m$) as the background set by annotating them as 0. $N$ denotes the number of spatial data points. $f_i(x^m)$ denotes the SDD of $x^m$ in semantic class $O_i$. We select the following uniform kernel function $K$:

$$K(x) = \mathbf{1}(x) = \begin{cases} 1, & |x| \leq 1 \\ 0, & otherwise \end{cases}. \quad (5)$$

***Definition.*** The meta-structures (MS) of a label are defined as a sematic class $O_i$, which is composed of pixels $x^m$ that have similar spatial density $f_i(x^m)$:

$$MS = \{O_1, O_2, \ldots\},$$

$$O_i = \{x^m \mid x^m \sim f_i(x^m)\} \ where \ (i = 1, 2, \ldots). \quad (6)$$

***Lemma 1***: If RCL/RL $Y^*$ is synthesized from $Y$ with $Q_{y^*|y}$, the $f_i(x^m)$ of $Y^*$ can be approximated as follows:

$$f_i(x^m) = \frac{1}{2Nh} \sum_{j=1}^{M} P(y^* = m | y = j) * S_j \pm \delta, \quad (7)$$

where $\delta$ denotes the sampling error, which is a constant, $S_j = W \cap O_j$ is the area in which $j_{th}$ semantic class $O_j$ overlaps with the search window $W$.

Because $N$, $h$, $\delta$ and $S_j$ are all constant, $f_i(x^m)$ is only dependent on the flipping probability $P(y^* = m | y = j; j = 1, 2, \ldots, M)$.

***Proof.*** By using kernel function $K$, $f_i(x^m)$ can be estimated by counting pixels within the search window:

$$f_i(x^m) = \frac{1}{2Nh} \sum_{k=1}^{N} \mathbf{1}(x - h \leq x_k^m \leq x + h). \quad (8)$$

Note that for RCL and RL, the number of counted pixels within the search window can be approximated by the flipping probability as follows:

$$\sum_{k=1}^{N} \mathbf{1}(x - h \leq x_k^m \leq x + h) = \sum_{j=1}^{M} P(y^* = m | y = j) * S_j \pm \delta, \quad (9)$$

Thus, the $f_i(x^m)$ can be approximated as follows:

$$f_i(x^m) = \frac{1}{2Nh} \sum_{j=1}^{M} P(y^* = m | y = j) * S_j \pm \delta. \quad (10)$$

$$QED$$

***Lemma 2***: If RCL/RL $Y^*$ is synthesized from $Y$ with $Q_{y^*|y}$, the number of semantic classes $D$ of $Y^*$ equals the rank $R$ of $Q_{y^*|y}$:

$$D = R. \quad (11)$$

***Proof.*** Based on *Lemma 1*, when the search window of pixel $x^m$ is within area $\{y = i\}$ (i.e., since the search window of near-boundary pixels may contain different semantic objects and these pixels are a minority, we can ignore these pixels for simplification), the $f_i(x^m)$ of semantic class $O_i$ equals:

$$f_i(x^m) = \frac{1}{2Nh} \sum_{j=1}^{M} P(y^* = m | y = j) * S_j \pm \delta$$

$$= \frac{1}{2Nh} * P(y^* = m | y = i) * 4h^2 \pm \delta$$

$$= \frac{2h}{N} * P(y^* = m | y = i) \pm \delta. \quad (12)$$

If rank $R$ of $Q_{y^*|y}$ is full, i.e., $R = M$, there exist at least $p$ and $q$ that $P(y^* = m | y = p) \neq P(y^* = m | y = q), \forall m$. Thus, the SDDs between arbitrary two classes are different, $f_p(x^m) \neq f_q(x^m), \forall p, q$. Based on the definition of meta-structures, the number of semantic classes $D = M = R$.

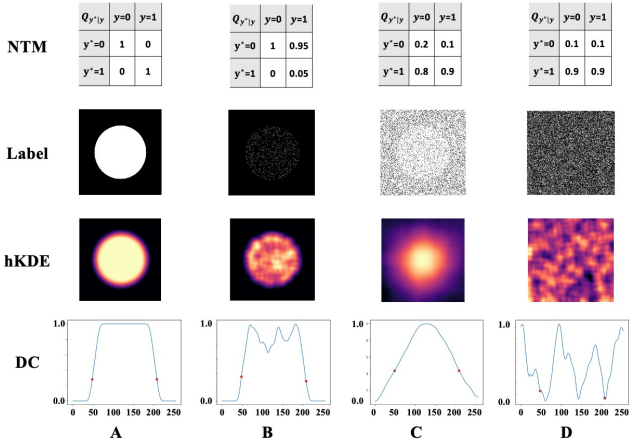

**Fig. 11** Spatial density distributions of binary-class semantic labels synthesized by different NTM. The heatmap of kernel density (hKDE) reveals the spatial density of each pixel label in the corresponding location. The density curve (DC) reveals values in the 128th row of hKDE, and the red points denote the location of object outlines in CL.

If rank $R$ of $Q_{y^*|y}$ is not full, i.e., $R < M$, there exits $(M - R + 1)$ linearly correlative columns in $Q_{y^*|y}$, indicating for corresponding semantic classes $O_{i,i=1,...,M-R+1}$, $P(y^* = m | y = i)$ equals a constant $P_m$. Thus, within the area $\{y = 1\} \cup \{y = 2\} \cup ... \cup \{y = M - R + 1\}$, the SDDs for these $(M - R + 1)$ classes are the same:

$$f_{i,i=1,...,M-R+1}(x^m) = \frac{2hP_m}{N} \pm \delta. \quad (13)$$

Based on the definition of meta-structures, the number of semantic classes $D = M - (M - R) = R$. In summary, $D = R$. *QED*

## 7.2 Quantifying Semantic Information in Labels

Based on *Lemma 1* and *Lemma 2*, we quantify the semantic information, referred to henceforth simply as semantics, of different labels using their SDDs.

**Semantics of CL and PCL.** Since CL and PCL contain all $M$ classes and their pixel labels are unrandomized, their semantics can be directly quantified by calculating SDD from kernel functions:

$$MS(CL) = \{O_1, O_2, ..., O_M\},$$

$$O_i = \{x^i \mid x^i \sim f_i(x^m)\} \text{ where } i = 1, 2, ..., M,$$

$$f_i(x^m) = \frac{1}{2Nh} \sum_{k=1}^{N} \mathbf{1}(x - h \le x_k^m \le x + h). \quad (14)$$

**Semantics of RCL.** For RCL, the NTM $Q_{y^*|y}$ can be arbitrary and the number of semantic classes equals the rank of $Q_{y^*|y}$, $D = R, R \in (1, M]$. Thus, the semantics of RCL can be quantified as:

$$MS(RCL) = \{O_1, O_2, ..., O_D\},$$

$$O_i = \{x^i \mid x^i \sim f_i(x^m)\} \text{ where } |i| = D = R \in (1, M],$$

$$f_i(x^m) = \frac{1}{2Nh} \sum_{j=1}^{M} P(y^* = m | y = j) * S_j \pm \delta. \quad (15)$$

**Semantics of RL.** For RL, the $Q_{y^*|y}$ can be arbitrary but its rank equals one. Thus, RL only contains one semantic class and can be quantified as:

$$MS(RL) = \{O_1\},$$

$$O_1 = \{x^1 \mid x^1 \sim f_1(x^m)\},$$

$$f_1(x^m) = \frac{1}{2Nh} \sum_{j=1}^{M} P(y^* = m | y = j) * S_j \pm \delta. \quad (16)$$

## 7.3 Theoretical Analysis of Learning Behavior of DNNs Trained by Noisy Labels

We present three theorems. *Theorem 1* demonstrates that RCL with full rank of $Q_{y^*|y}$ contains the same semantics as CL. This explains why DNNs trained by RCL have similar performance as DNNs trained CL. *Theorem 2* demonstrates that semantics of PCL are changed in the perturbation area. This explains why PCL degrades performance of DNNs in both feature extraction and segmentation. *Theorem 3* demonstrates that semantics in labels is more dependent on the rank of the noise transition matrix. This further explain why DNNs learn semantic segmentation from meta-structures rather than pixel labels per se. Detailed proofs are given below.

**Theorem 1:** If RCL/RL $Y^*$ is synthesized from $Y$ with full rank of $Q_{y^*|y}$, the noisy labels $Y^*$ have the same semantics as CL $Y$.

**Proof.** Based on *Lemma 2*, if the rank of $Q_{y^*|y}$ equals $M$, the meta-structures of $Y^*$ and the meta-structures of $Y$ have the same number of semantic classes:

$$|MS(Y^*)| = |MS(Y)| = |\{O_1, O_2, ..., O_M\}| = M. \quad (17)$$

Because the boundaries of semantic objects in $Y^*$ are not changed when synthesized from $Y$, the locations of differences between different SDDs are also unchanged, indicating that each semantic area in $Y^*$ is consisted by the same pixels of the corresponding semantic area in $Y$. Thus, the meta-structures of $Y^*$ is identical to the meta-structures of $Y$, and they have the same semantics. *QED*

We also demonstrate *Theorem 1* by experiments. We simulate a simplified binary-class CL ($256 \times 256$ image) of that has a circle within a rectangle as shown in Fig. 11A. We synthesize RCLs with different $Q_{y^*|y}$ as shown in Fig. 11B-C and RLs as shown in Fig. 11D.

The SDD of labels are visualized in two ways: heatmaps of kernel density estimation (hKDE) and density curves (DC). The hKDE directly shows spatial density of each pixel label at its corresponding location. The DC shows the spatial density of each pixel in the 128th row of the image, and the red points denote the outline location of the semantic class in CL.

Figure 11A shows the hKDE and DC of the CL. The hKDE clearly shows two different SDDs and the high-density values are aggregated in the circular region (foreground). Similar as shown the RL whose rank of $Q_{y^*|y}$ equals 1 (Figure 11D), the hKDE and DC are randomized. And we cannot distinguish between two classes from their SDDs, indicating the loss of semantics because the two classes are fused into one.

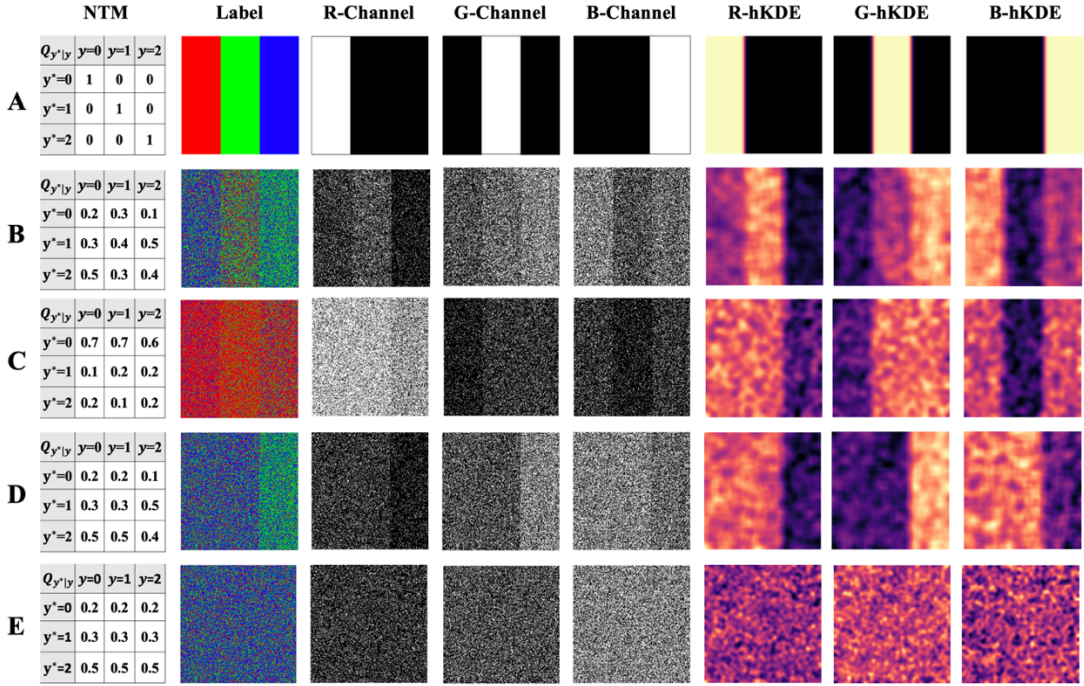

**Fig.12** Spatial density distributions of three-class labels synthesized by different NTM. For each class channel, the heatmap of kernel density (hKDE) reveals the spatial density of each pixel label in the corresponding location.

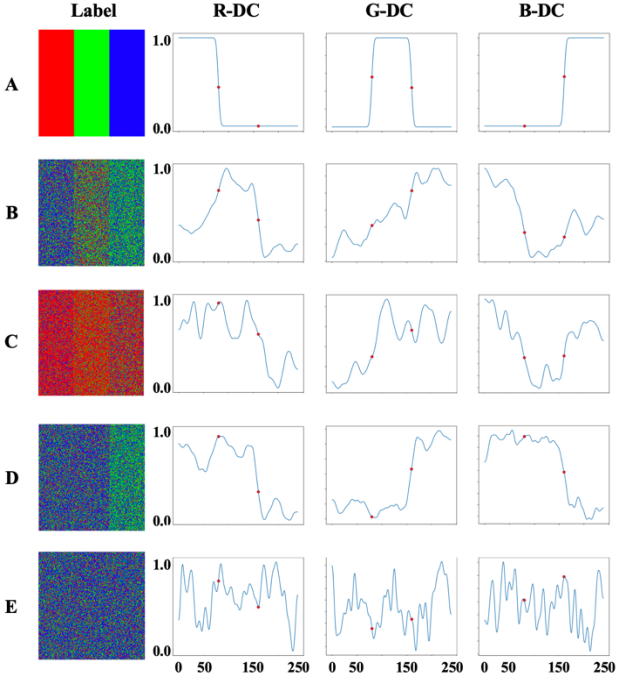

**Fig.13** Density curves (DC) of three-class labels in each class channels. The red points denote the location of object outlines in CL.

**Table 8** Comparison Results of DNNs Trained by CL and Dynamic RCL

| Model | Dataset | CL | Dynamic RCL |
|---|---|---|---|
| U-Net | ER | 75.4 | 74.8 |
| | MITO | 67.8 | 67.1 |
| DeepLabv3+ | Cityscapes | 64.8 | 64.2 |
| | Cityscapes-3 | 94.3 | 93.9 |

We have obtained consistent results on multi-class semantic labels after dividing labels into multiple binary-class labels. The results of hKDE and DC are shown in Figure 12 and Figure 13, respectively. As we can see, with the decreasing of the rank in NTM, the SDD difference between corresponding classes disappears.

**Theorem 2:** For PCL $Y^*$, the more unrandomized pixel perturbed on CL $Y$, the more perturbation on its semantics.

**Proof.** The semantic area $A$ of a class in PCL can be divided into an unperturbed area $A'$ and a set of perturbed area $P_i$ that one class labels within $P_i$ are all flipped to another class:

$$A = A' \cup \{P_i, i = 1,2, \dots\}. \tag{18}$$

Within $A'$, the $Q_{y^*|y}$ is an identity matrix. Within $P_i$, the $Q_{y^*|y}$

in DC, the spatial density of one part of pixels is near 1 and the others are near 0. Both hKDE and DC reveal that meta-structures of CL contain two semantic classes. Each of them consists of pixels that have similar spatial density distributions.

For the RCL whose rank of $Q_{y^*|y}$ equals 2 (Figure 11B-C), all of their hKDE and DC show two different SDDs that are similar to CL, indicating that RCL has similar meta-structures as CL. Thus, RCL has similar semantic information as CL. However, for contains two identical columns, indicating the rank decreases in $Q_{y^*|y}$. Examples of $Q_{y^*|y}$ for two-class PCL are shown in Figure 2. Based on *Lemma 2*, semantics does not change in area $A'$ but does change in area $P_i$. Meanwhile, more perturbed pixels on CL will increase the area of $P_i$, indicating more perturbation on semantics. *QED*

**Theorem 3:** If DNNs are trained by *dynamic* full-rank RCL, which means RCL is synthesized by different full-rank $Q_{y^*|y}$ under different epochs, and DNNs are optimized by empirical risk minimization (ERM) (Vapnik 1992), it will have similar performance as DNNs trained by CL.

**Proof:** If we view $Q_{y^*|y}$ as a random variable $Q$, the expected risk $R(f)$ of DNNs trained by dynamic RCL is given by:

$$R(f) = \iint \mathcal{L}(f(x), y) \, dP(x, y) dP(Q), \quad (19)$$

where $f$ is the learning system, and $L$ is the loss function. Since distributions $P(x, y)$ and $P(Q)$ are unknown, we can use empirical risk to approximate $R(f)$:

$$R(f) = \frac{1}{TN} \sum_{j=1}^{T} \sum_{i=1}^{N} \mathcal{L}(f(x_i), y_i). \quad (20)$$

where $N$ is the number of training samples, $T$ is the number of $Q_{y^*|y}$ that equals the number of training epochs. If $f_j = argmin \frac{1}{N} \sum_{i=1}^{N} \mathcal{L}(f(x_i), y_i)$, we have:

$$f^* = argmin \, R(f) = \frac{1}{N} \sum_{j=1}^{N} f_j. \quad (21)$$

This indicates that $f^*$ optimized by dynamic RCL can be viewed as the boosting of a series of $f_j$, where each $f_j$ is trained by RCL with the different full-rank $Q_{y^*|y}$.

Since DNNs trained by RCL with full-rank $Q_{y^*|y}$ have similar performance as DNNs trained by CL (*Theorem 1*), the DNNs trained by dynamic RCL also have similar performance as CL. Experimental demonstration of *Theorem 3* is shown in Table 8. *QED*

## 7  Conclusion

In this study, we examine the learning behavior of DNNs trained by different types of pixel-level noisy labels in semantic segmentation and provide direct experimental evidence and theoretical proof that DNNs learn meta-structures from noisy labels. The unsupervised segmentation model we have developed provides an example on how to utilize the meta-structures in practice. However, our study also has its limitations. In particular, our model of meta-structures remains to be further developed, and utilization of meta-structures remains to be extended to other applications such as multi-class segmentation. Despite these limitations, the learning behavior of DNNs revealed in this study provides new insight into what and how DNNs learn from noisy labels to segment images.

## Acknowledgment

The authors thank colleagues at CBMI for their technical assistance. The work was supported in part by the National Natural Science Foundation of China (grant 31971289 and 91954201 to G.Y.) and the Strategic Priority Research Program of the Chinese Academy of Sciences (grant XDB37040402 to G.Y.).

## Statements and Declarations

### Competing Interests

The authors declare no competing interests.

### Data Availability

The Cityscape (Cordts et al. 2016) dataset is available in https://www.cityscapes-dataset.com/.

The Cityscape-3 dataset is generated by masking the irrelated classes from the Cityscape dataset, which is available in https://www.cityscapes-dataset.com/.

The endoplasmic reticulum (ER) and mitochondria (MITO) (Luo et al. 2020) datasets are available in https://ieee-data-port.org/documents/fluorescence-microscopy-image-datasets-deep-learning-segmentation-intracellular-orgenelle.

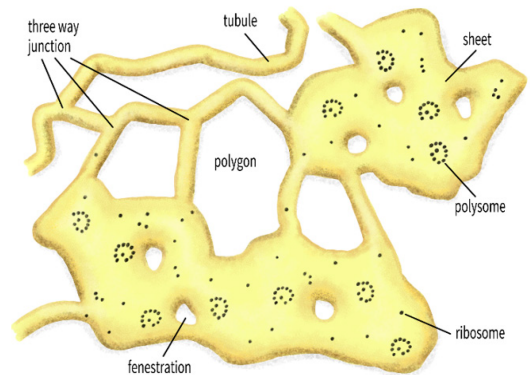

**Fig.14** An illustration of ER network of sheets and tubules (Sree et al. 2021). Network polygons and branch points (called three-way junctions), ribosomes, polysomes, and fenestrations on sheets are indicated.

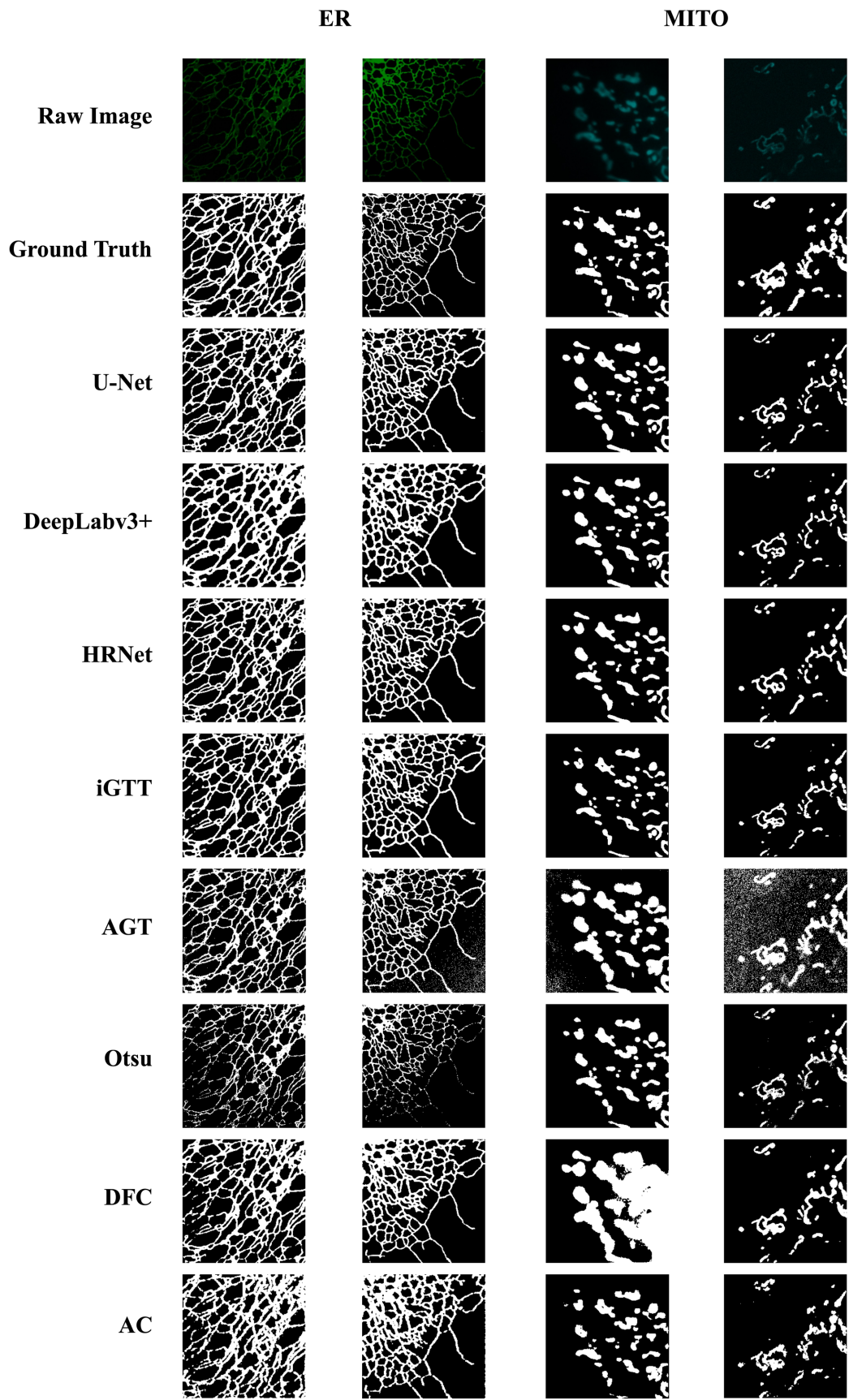

**Fig.15** Visualization results of different models on ER, MITO datasets.